\documentclass[10pt,twocolumn,letterpaper]{article}
\usepackage[pagenumbers]{wacv} 

\usepackage{graphicx}
\usepackage{amsmath}
\usepackage{amssymb}
\usepackage{booktabs}
\usepackage{algorithm}
\usepackage{algorithmic}
\usepackage{mathrsfs}
\usepackage{xcolor}
\usepackage{multirow}
\usepackage{colortbl}
\usepackage{epsfig}
\definecolor{cjh}{rgb}{0,0.8,0.6}
\definecolor{mygray}{gray}{.88}
\usepackage[pagebackref,breaklinks,colorlinks]{hyperref}

\usepackage[capitalize]{cleveref}
\crefname{section}{Sec.}{Secs.}
\Crefname{section}{Section}{Sections}
\Crefname{table}{Table}{Tables}
\crefname{table}{Tab.}{Tabs.}

\usepackage[accsupp]{axessibility}  

\begin{document}

\title{Spiking Denoising Diffusion Probabilistic Models}

\author{
    Jiahang Cao\textsuperscript{\rm 1}\thanks{Equal contribution.} ~~ 
    Ziqing Wang\textsuperscript{\rm 1,2}\footnotemark[1] ~~
    Hanzhong Guo\textsuperscript{\rm 3}\footnotemark[1] ~~
    Hao Cheng\textsuperscript{\rm 1}~
    Qiang Zhang\textsuperscript{\rm 1}~
    Renjing Xu\textsuperscript{\rm 1}\thanks{Corresponding author.} \\
    \textsuperscript{\rm 1}\small{The Hong Kong University of Science and Technology (Guangzhou)}\\
    \textsuperscript{\rm 2}\small{North Carolina State University,}
    \textsuperscript{\rm 3}\small{Renmin University of China}\\
    {\tt\small \{jcao248,hcheng046,qzhang749\}@connect.hkust-gz.edu.cn,}\\
 {\tt\small guohanzhong@ruc.edu.cn, zwang247@ncsu.edu, renjingxu@hkust-gz.edu.cn}
}
\maketitle

\begin{abstract}
    Spiking neural networks (SNNs) have ultra-low energy consumption and high biological plausibility due to their binary and bio-driven nature compared with artificial neural networks (ANNs). While previous research has primarily focused on enhancing the performance of SNNs in classification tasks, the generative potential of SNNs remains relatively unexplored. In our paper, we put forward Spiking Denoising Diffusion Probabilistic Models (SDDPM), a new class of SNN-based generative models that achieve high sample quality. To fully exploit the energy efficiency of SNNs, we propose a purely Spiking U-Net architecture, which achieves comparable performance to its ANN counterpart using only 4 time steps, resulting in significantly reduced energy consumption. Extensive experimental results reveal that our approach achieves state-of-the-art on the generative tasks and substantially outperforms other SNN-based generative models, achieving up to $12\times$ and $6\times$ improvement on the CIFAR-10 and the CelebA datasets, respectively. Moreover, we propose a threshold-guided strategy that can further improve the performances by 2.69\% in a training-free manner. The SDDPM symbolizes a significant advancement in the field of SNN generation, injecting new perspectives and potential avenues of exploration. Our code is available at 
    \href{https://github.com/AndyCao1125/SDDPM}{https://github.com/AndyCao1125/SDDPM}.
\end{abstract}
\vspace{-10pt}
\section{Introduction}

Spiking neural networks (SNNs), being regarded as the third generation of neural networks, are potential competitors to artificial neural networks (ANNs) due to their distinguished properties: high biological plausibility, event-driven nature, and low power consumption. In SNNs, all information is represented by binary time series data rather than continuous representation, which allows SNNs to adopt low-power accumulation (AC) instead of the traditional high-power multiply-accumulation (MAC), leading to significant energy efficiency gains. Existing works reveal that on specialized hardware, such as Loihi~\cite{davies2018loihi} and TrueNorth~\cite{akopyan2015truenorth}, SNNs can save energy by orders of magnitude compared with ANNs.
Additionally, SNNs follow their biological counterparts and inherit complex temporal dynamics from them, endowing SNNs with powerful abilities to extract spatial-temporal features in a variety of tasks, including recognition~\cite{wang2023masked,zhou2022spikformer,deng2022temporal}, tracking~\cite{zhang2022spiking}, segmentation~\cite{kirkland2020spikeseg} and images restoration~\cite{li2021event}.

\begin{figure}[t]
\begin{center}
\vspace{3pt}
\begin{tabular}{@{}c@{}}
\includegraphics[width=0.78\linewidth]{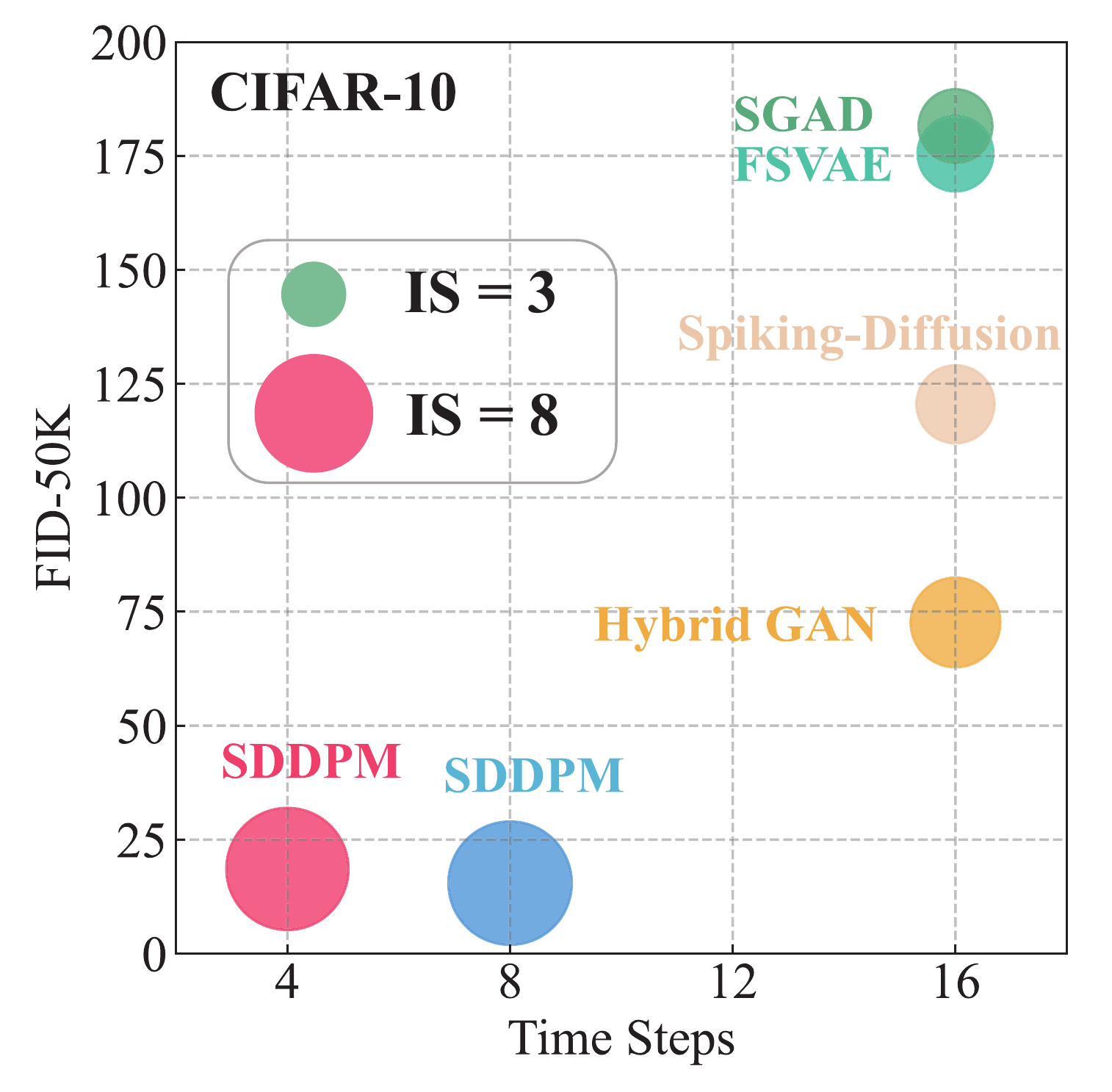} \vspace{-3mm} \\
\end{tabular}
\vspace{-5mm}
\end{center}
\caption{\textbf{Comparisons of the SNN-based generative models.} The Fr\'{e}chet Inception Distance (FID) serves as a measure of image quality, with lower values indicating superior performance. The Inception Score (IS) acts as an alternate measure of model performance, where a higher score is desirable. The size of the markers denotes the IS score. In comparison to other SNN-based generative models, our models demonstrate state-of-the-art performance with fewer time steps.}
\vspace{-5mm}
\label{fig:compare_start}
\end{figure}

However, most of the existing research on SNNs primarily focuses on classification-based tasks, and the regression capability of SNNs has not been well demonstrated, especially in image generation tasks. 
Spiking-GAN~\cite{kotariya2022spiking} is the first SNN-based image generation model, but its low performance and limited experimentation on handwritten data hinder a sufficient demonstration of its generative ability on high-dimensional data. Kamata \etal~\cite{kamata2022fully} propose a fully spiking variational autoencoder (FSVAE) combined with discrete Bernoulli sampling and claim that the quality of the generated images surpasses the ANN-based VAE in the same setting but the quality of the generated images is limited for practical applications.~\cite{ho2020denoising,nichol2021improved}. Consequently, it is imperative to develop a generative algorithm capable of producing high-quality images while also reducing energy consumption.

Recently, diffusion models have achieved remarkable success in generation tasks~\cite{dhariwal2021diffusion,karras2022elucidating} since they offer several advantages compared with other deep generative models (DGMs). Firstly, diffusion's regression loss makes its training more stable than the adversarial loss in the Generative Adversarial Networks (GANs) and therefore the diffusion model is more suitable for large-scale generation tasks.~\cite{rombach2022high}. Secondly, the training objective of diffusion models is derived directly from the likelihood perspective, so the problem of mode collapse can be avoided when the model converges. 
Furthermore, since the diffusion models can be viewed as a VAE for a given encoder, it is easier to optimize. 
These advantages provide the impetus for us to investigate the feasibility of incorporating SNNs into the diffusion model, leveraging the generative capabilities of diffusion models along with the energy efficiency inherent in SNNs.

In this work, we propose Spiking Denoising Diffusion Probabilistic Models (SDDPM), a novel category of SNN-based diffusion models exhibiting exceptional image generation capabilities. 
To fully leverage the energy efficiency of SNNs, we propose the Spiking U-Net architecture that achieves comparable performance to its ANN counterpart while employing only 4 spiking time steps, resulting in significantly reduced energy consumption. 
Moreover, we employ a pre-spike structure to ensure the accurate transmission of spikes. 
We also propose training-free threshold guidance, which further enhances the quality of the generated images by adjusting the threshold value of the spiking neurons.
Comprehensive experimental results demonstrate that 
threshold guidances contribute to the facilitation of SDDPM. Our approach is evaluated on four datasets: MNIST, Fashion-MNIST, CIFAR-10, and CelebA.  
As shown in Fig.~\ref{fig:compare_start}, we demonstrate that the proposed SDDPM outperforms all SNN-based generative models by a significant margin, requiring only a small number of spiking time steps. We also conduct extensive ablation studies to reveal the effectiveness of each component. To sum up, our contributions lie in four folds:
\begin{itemize}
    \item To the best of our knowledge, SDDPM is the first work that employs spiking neural networks on diffusion models.
    \item To fully exploit the energy efficiency of SNN, We design a purely Spiking U-Net that can achieve comparable performance to its ANN counterpart while saving 62.5\% energy consumption.
    \item Extensive experiments show that SDDPM achieves state-of-the-art performances among SNN-based generative models. Specifically, our proposed SDDPM outperforms the SNN-based baselines by up to 1200\% and 600\% on the CIFAR-10 and CelebA datasets with only 4 spiking time steps. 
    \item  We also introduce a threshold-guidance strategy aimed at further enhancing performance, which results in a 2.69\% improvement without any additional training.
\end{itemize}

\begin{figure*}[t!]
	\setlength{\tabcolsep}{1.0pt}
	\centering
	\begin{tabular}{c}
		\includegraphics[width=1\textwidth]{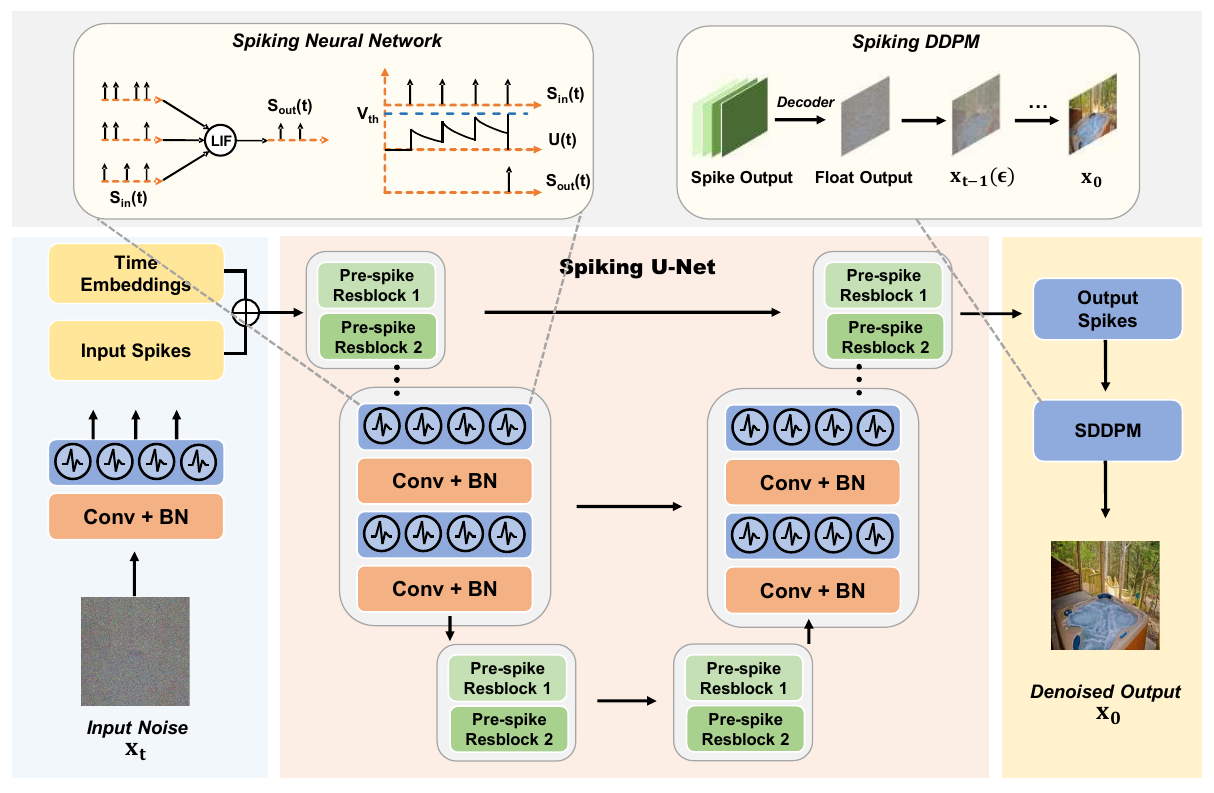} 
	\end{tabular}
        \vspace{-0.35cm}
	\caption{\textbf{Illustration of the architecture and pipeline of Spiking Diffusion Models.} The SDDPM architecture is suitable for use on top of any existing diffusion models, where we inherit the most commonly used U-Net backbone and propose the Spiking U-Net. Our network consists of several Pre-spike Resblocks (colored in green), each of which contains spiking neurons (blue) and Conv-BatchNorm layers (orange). Given a random Gaussian noise input $x_t$, it is converted into the spikes by an encoding layer and subsequently fed into the Spiking U-Net along with the time embeddings. The network transmits only spikes, represented by $0/1$ vector. The output spikes $S_{out}(t)$ are formed as a result of the accumulation of membrane potentials $U(t)$ within the neuron under the influence of the consecutive input spikes $S_{in}(t)$. Once the membrane potential exceeds the threshold $V_{th}$, the neuron will generate a spike. Eventually, the output spikes are passed through a decoding layer to obtain the predicted noise $\epsilon$, followed by $N$ times denoising to restore the image $x_0$.
 }
	\label{fig:pipeline}
	\vspace{-0.5cm}
\end{figure*}

\section{Related Work}

\noindent\textbf{Training Methods of Spiking Neural Networks.}
Generally, there are two ways to obtain deep SNN models: ANN-to-SNN conversion and direct training. The ANN-to-SNN conversion~\cite{bu2023optimal,deng2021optimal,ding2021optimal,li2021free,wang2023masked} involves converting a pre-trained ANN into an SNN by replacing the ReLU activation layers with spiking neurons, which allows the SNN to simulate the behavior of the original ANN using spiking neurons. 
This conversion method is generally known to achieve higher accuracy compared to direct training methods. 
However, the conversion methods typically require a longer training time compared to direct training methods, resulting in the need for more training resources.
On the other hand, direct training methods involve training the SNN directly from scratch. 
{Surrogate gradients~\cite{neftci2019surrogate, lee2020enabling, xiao2021training} are utilized for addressing the non-differentiability problem of spiking neurons, enabling the training of SNNs using gradient-based optimization techniques.}
In our study, we explore the feasibility of implementing diffusion models in SNNs using the direct train method, aiming to reduce power consumption and investigate the potential generative abilities of SNNs.

\vspace{3pt}\noindent{\textbf{Diffusion Models.}
Some research focuses on analyzing the theoretical foundation and formulation of diffusion models~\cite{ho2020denoising,song2020denoising,song2019generative,song2020score}. What's more, diffusion model is divided into discrete time diffusion and continuous time diffusion~\cite{song2020score,karras2022elucidating,kingma2021variational} depending on whether the time step in diffusion is sampled from discrete distribution or continuous distribution. Certain solvers~\cite{jolicoeur2021gotta,bao2022analytic,dockhorn2022genie,lu2022dpm} have been proposed to expedite the sampling process in diffusion models. Additionally, some studies are dedicated to designing more efficient diffusion networks~\cite{peebles2022scalable,bao2023all}.}

\vspace{3pt}\noindent\textbf{Spiking Neural Network in Generative Models.}
There has been some prior research investigating the capabilities of SNNs in generative tasks.
VDIB~\cite{skatchkovsky2021learning} is a hybrid variational autoencoder, consisting of an SNN-based encoder and an ANN-based encoder. 
Hybrid guided-VAE~\cite{stewart2022encoding} and hybrid GAN~\cite{rosenfeld2022spiking} also adopt SNN-ANN architecture.
However, the aforementioned approaches rely on the ANNs, resulting in the entire model not being fully deployed on neuromorphic hardware.
Spiking GAN~\cite{kotariya2022spiking} incorporates a fully SNN-based backbone and utilizes a time-to-first-spike coding scheme. 
Kamata \etal~\cite{kamata2022fully} introduce a fully spiking variational autoencoder (FSVAE), which samples images according to the Bernoulli distribution.
Recently, Feng \etal~\cite{feng2023sgad} construct a spiking generative adversarial network with attention-scoring decoding for handling complex images, and Liu \etal~\cite{liu2023spiking} propose a spike-based vector quantized variational autoencoder (VQ-SVAE) to learn a discrete latent space for images.
However, the primary limitation of existing spiking generative models is their low performance and poor generated image quality. These drawbacks hinder their competitiveness in the field of generative models, despite their low energy consumption. To tackle this issue, we introduce the Spiking Denoising Diffusion Probabilistic Model (SDDPM), which not only delivers substantial improvements over existing SNN-based generative models but also preserves the advantages of SNNs.

\section{Background}

\subsection{Spiking Neural Network}
The spiking neural network is a bio-inspired algorithm that mimics the actual signaling process occurring in brains.
Compared to the artificial neural network, it transmits sparse spikes instead of continuous representations, offering benefits such as low energy consumption and robustness. 
In this paper, we adopt the widely used Leaky Integrate-and-Fire (LIF~\cite{hunsberger2015spiking,burkitt2006review}) model, which effectively characterizes the dynamic process of spike generation and can be defined as:
\begin{equation}
    \tau\frac{\mathrm{d} V(t)}{\mathrm{d} t}= - (V(t) - V_{\textrm{reset}}) + I(t),
\end{equation}
where $I(t)$ represents the input synaptic current at time $t$ to charge up to produce a membrane potential 
  $V(t)$, $\tau$ is the time constant. When the membrane potential exceeds the threshold $\vartheta_{\textrm{th}}$, the neuron will trigger a spike and resets its membrane potential to a value $V_{\textrm{reset}}$ ($V_{\textrm{reset}}<\vartheta_{\textrm{th}}$). The LIF neuron achieves a balance between computing cost and biological plausibility. 

In practice, the dynamics need to be discretized to facilitate reasoning and training. The discretized version of LIF model can be described as:
\begin{align}
    & U[n] = e^{\frac{1}{\tau}}V[n-1] + I[n] \label{eq:dis_lif1},\\
    & S[n] = \Theta (U[n] - \vartheta_{\textrm{th}})\label{eq:dis_lif2},\\
    & V[n] = U[n](1-S[n]) + V_{\textrm{reset}}S[n] \label{eq:dis_lif3},
\end{align}
where $n$ is the time step, $U[n]$ is the membrane potential before reset, $S[n]$ denotes the output spike which equals 1 when there is a spike and 0 otherwise, $\Theta(x)$ is the Heaviside step function, $V[n]$ represents the membrane potential after triggering a spike. In addition, we use the “hard reset” method~\cite{fang2021incorporating} for resetting the membrane potential in Eq.~\eqref{eq:dis_lif3}, which means that the value of the membrane potential $V[n]$ after triggering a spike ($S[n]=1$) will go back to $V_{\textrm{reset}} = 0$. 

\subsection{Diffusion Models and Classifier Guidance}
Diffusion models gradually perturb data with a forward diffusion process and then learn to reverse such process to recover the data distribution. 

Formally, let $x_0\in\mathbb{R}^n$ be a random variable with unknown data distribution $q(x_0)$. The forward diffusion process $\left \{ x_t \right \} _{t\in [0,T]}$ indexed by time $t$, can be represented by the following forward stochastic differential equations~(SDE):
\begin{eqnarray}
\label{eq:sdef}
     \textrm{d} x_t = f(t)x_t\textrm{d} t + g(t)\textrm{d} \omega  ,\quad x_0\sim q(x_0),
\end{eqnarray}
where $\omega \in \mathbb{R}^n$ is a standard Wiener process. 
Let $q(x_t)$ be the marginal distribution of the above SDE at time $t$. Its corresponding reversal process can be described by another SDE which recovers the data distribution from noise~\cite{song2020score}:
\begin{eqnarray}
\label{eq:continuous_back}
    \textrm{d} x = \left [ f(t)x_t-g^2(t) \nabla_{x_t} \log q(x_t) \right ] \textrm{d} t + g(t)\textrm{d} \bar{\omega} ,
\end{eqnarray}
where $\bar{\omega} \in \mathbb{R}^n$ is a reverse-time standard Wiener process and this reversal SDE starts from $x_T\sim q(x_T)$. 
In Eq.~\eqref{eq:continuous_back}, the only unknown term is the score function $\nabla_{x_t} \log q(x_t)$. To estimate this term, prior works\cite{ho2020denoising, song2020score, karras2022elucidating} employ a noise network $\epsilon_{\theta}(x_t,t)$ to estimate scaled score function $\sigma(t)\nabla_{x_t} \log q(x_t)$ via denoising score matching (DSM)~\cite{vincent2011connection}, which ensures that the optimal solution satisfies $\epsilon_{\theta}(x_t,t)=-\sigma(t) \nabla_{x_t} \log q(x_t)$, where $\sigma(t)$ denotes the variance of $q(x_t|x_0) \sim \mathcal{N}(x_t|a(t)x_0,\sigma^2(t)I)$, which is related to the notation in Eq.~\eqref{eq:sdef} as shown in Eq.~\eqref{eq:gab},
\begin{eqnarray}
\label{eq:gab}
    f(t)=\frac{\textrm{d} \log a(t)}{\textrm{d} t} ,\quad g^2(t)=\frac{\textrm{d} \sigma^2(t)}{\textrm{d} t}-2\sigma^2(t)\frac{\textrm{d} \log a(t)}{\textrm{d} t} .
\end{eqnarray}
Hence, sampling can be achieved by discretizing the reverse SDE in
Eq.~\eqref{eq:continuous_back} by replacing the $\nabla_{x_t} \log q(x_t)$ with noise network $-\frac{\epsilon_{\theta}(x_t,t)}{\sigma(t)}$.
Furthermore, to enable conditional sampling, such as sampling cat images, we can refine the reverse stochastic differential equation (SDE) presented in Eq.~\eqref{eq:continuous_back} as follows~\cite{dhariwal2021diffusion}:
\begin{eqnarray}
\label{eq:condition}
 & \epsilon _{\theta}(x_t,c) = \epsilon _{\theta}(x_t)-s \sigma(t)\nabla_{x_t}\log p_{\phi}(c|x_t,t),
\end{eqnarray}
Here, $p_{\phi}(c|x_t,t)$ represents the classifier, $s$ denotes the temperature controlling the intensity of guidance, and Eq.~\eqref{eq:condition} indicates that a conditional sample can be generated using only a pre-trained noise network and a classifier. 
Ho \etal~\cite{ho2022classifier} introduced classifier-free guidance, which significantly enhances the diversity of generated samples. This methodology has found extensive application in practical scenarios~\cite{IF}, as demonstrated by the works of Ho .~\cite{ho2022classifier}.

Furthermore, it is important to note that the guidance mentioned above is not limited to a specific category, which can be applied to various forms of guidance. For example, in some studies~\cite{zhao2022egsde, bao2022equivariant}, energy-based guidance is proposed to facilitate image translation and molecular design. Additionally, Kim \etal~\cite{kim2022refining} introduce discriminator guidance to mitigate estimation bias of the noise network, resulting in state-of-the-art performance on the CIFAR-10 dataset.

\section{Method}
In this section, we introduce our methodologies in three stages. In Sec.~\ref{subsec:sunet}, we introduce our proposed Spiking U-Net and provide a comprehensive explanation of its network architecture. Then, we present the pre-spike residual structure in Sec.~\ref{subsec:pre_spike}. 
Eventually, we put forward a threshold-guiding strategy and its corresponding theory in Sec.~\ref{subsec:tg}. The computational formulations for calculating the energy consumption of the SNNs are given in the Supplementary Material. 

\subsection{Spiking U-Net Structure}
\label{subsec:sunet}
 The overview of the architecture and sampling pipeline is illustrated in Fig.~\ref{fig:pipeline}. Spiking U-Net is the main component of the whole SDDPM structure. Unlike previous work~\cite{skatchkovsky2021learning, stewart2022encoding, rosenfeld2022spiking} that use hybrid architecture consisting of SNN and ANN, we introduce a purely SNN-based structure, thereby fully leveraging the enhanced energy efficiency inherent to SNNs.

The ANN-based U-Net utilized in DDPM~\cite{ho2020denoising} is characterized by a residual block (resblock) defined as:
\begin{equation}
O^l = Conv^l(Swish(GN^l(O^{l-1}))) + O^{l-1},
\end{equation}
where $O^l$ is the output representation at layer $l$, $GN$ signifies the group normalization operation, and $Swish$~\cite{ramachandran2017searching} represents the activation function.

However, directly employing $GN$ in SNNs may result in performance degradation due to distribution mismatch~\cite{wang2023masked}. Consequently, we substitute the $GN$ in the U-Net architecture with batch normalization, which is a more SNN-compatible normalization technique~\cite{fang2021deep, zhou2022spikformer}. This modification allows the model to better capture spatial features. 
The residual block in our Spiking U-Net can be formulated as follows:
\begin{align}
    O^l &= BN^l(Conv^l(S^{l-1})) + S^{l-1},\\
    S^l &= SpikeNeuron(O^{l}),\\
    O^{l+1} &= BN^{l+1}(Conv^{l+1}(S^{l})) + S^{l},\\
    S^{l+1} &= SpikeNeuron(O^{l+1}),
\end{align}
where $S^l$ is the output spikes at layer $l$, $BN$ denotes the batch normaliztion operation and $SpikeNeuron$ means the spiking activation function in Eq.~\eqref{eq:dis_lif2}.

The Spiking U-Net receives an input of a 2D image batch $I_s \in \mathbb{R}^{B \times C \times H \times W}$, with $B, C, H$, and $W$ standing for batch size, channel, height, and width, respectively. Initially, the image is replicated $T$ times, resulting in a sequence of images $I \in \mathbb{R}^{T \times B \times C \times H \times W}$, a necessary operation for the SNN to incorporate temporal dimension information. However, the 2D convolution and BN cannot directly process the added $T$ dimension. To circumvent this, we fuse the $T$ and $B$ dimensions, represented mathematically as $I_{\text{fused}} \in \mathbb{R}^{TB \times C \times H \times W}$, which allows the network to concurrently analyze spatial and temporal features.

\subsection{Pre-spike Residual Learning}
\label{subsec:pre_spike}

In this section, we further explore the structure of the Spiking U-Net. Although the above design can fully apply the U-Net into SNN, it could cause the output range of the residual block to overflow. This is due to the fact that the previous shallow network output $S^{l-1}$ and the residual mapping representation $S^l$ are both spike series ($\{0,1\}$), thus their summation $O^l$ would result in a value domain of $\{0,1,2\}$, where $\{2\}$ is a pathological case without any biological plausibility. 
{This could lead to a larger range of spike signals when the layers become deeper~\cite{zhou2023spikingformer}, incurring higher energy consumption.}   

\begin{figure}[t]
\begin{center}
\vspace{3pt}
\begin{tabular}{@{}c@{}}
\includegraphics[width = 0.95\linewidth]{{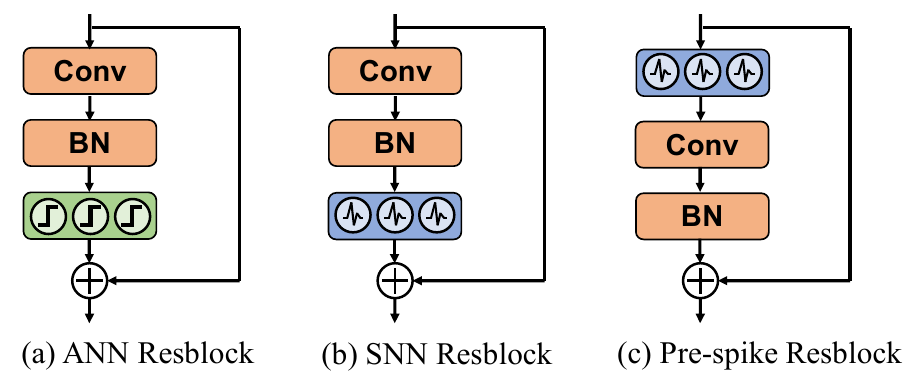}} \vspace{-3mm} \\
\end{tabular}
\vspace{-3mm}
\end{center}
\caption{\textbf{Comparisons of the residual structures and Pre-spike structure.} Standard SNN resblock (b) entirely inherits from ANN structure (a). In contrast, pre-spike resblock activates first. }
\vspace{-2mm}
\label{fig:resblock}
\end{figure}

Inspired by~\cite{liu2018bi,zhang2022pokebnn}
, we for the first time apply pre-spike residual learning with the structure of $Activation$-$Conv$-$BatchNorm$
in our Spiking U-Net, so as to overcome the problem of gradient explosion/vanishing
and performance degradation in convolution-based SNNs. Through the pre-spike
blocks, the residuals, and outputs are summed by floating point addition operation, ensuring that the representation is accurate before entering the next spiking neuron while avoiding the pathological condition mentioned above. The whole pre-spike residual learning process inside a resblock can be formulated as below:
\begin{align}
    S^{l} &= SpikeNeuron(O^{l-1}),\\
    O^{l} &= BN^l(Conv^l(S^l)) + O^{l-1},\\
    S^{l+1} &= SpikeNeuron(O^{l}),\\
    O^{l+1} &= BN^{l+1}(Conv^{l+1}(S^{l+1})) + O^{l}.
\end{align}
Through the pre-spike residual mechanism, the output of the residual block can be summed by two floating points $BN^l(Conv^l(S^l))$, $O^{l-1}$ at the same scale and then enter the spiking neuron at the beginning of the next block, which guarantees that the energy consumption is still very low. We illustrate the diagram of different resblocks in Fig.~\ref{fig:resblock}. Experiments to verify the superiority of the pre-spike structure can be found in Sec.~\ref{subsec:ablation_study}.

\begin{table*}[t]
\setlength\tabcolsep{12pt} 
\centering
\resizebox{1\linewidth}{!}{
\begin{tabular}{clccccc}
\toprule 
\multirow{1}{*}{\textbf{Dataset}} &
  \multicolumn{1}{c}{\textbf{Model}}&
  \multirow{1}{*}{\textbf{Method}} &
  \multirow{1}{*}{\textbf{\#Param (M)}} &
  \multirow{1}{*}{\textbf{Time Steps}} &
  \multirow{1}{*}{\textbf{IS}$\uparrow$} &
  \multirow{1}{*}{\textbf{FID}$\downarrow$} \\ \midrule
\multirow{5}{*}{MNIST$^*$} 
& {DDPM}~\cite{ho2020denoising} & {ANN} & {64.47} & {/} & {-} & {28.70}\\
\cmidrule{2-7}
& \multicolumn{1}{l}{FSVAE~\cite{kamata2022fully}} &SNN &3.87 & 16 & 6.209 & 97.06        \\
& \multicolumn{1}{l}{SGAD~\cite{feng2023sgad}} &SNN & -& 16 & - & 69.64 \\
& \multicolumn{1}{l}{{Spiking-Diffusion~\cite{liu2023spiking}}} &{SNN} & {-}& {16} & {-} & {37.50} \\       
& \multicolumn{1}{l}{\textbf{SDDPM} }&SNN& 63.61 & 4 & - & \textbf{29.48}\\\midrule
\multirow{5}{*}{\begin{tabular}[c]{@{}c@{}}Fashion\\ MNIST$^*$\end{tabular}} 
& {DDPM}~\cite{ho2020denoising} & {ANN} & {64.47} & 
{/} & {-} & {20.24}\\
\cmidrule{2-7}
& FSVAE~\cite{kamata2022fully} &SNN & 3.87 & 16 &  4.551 & 90.12  \\
& SGAD~\cite{feng2023sgad} &SNN &-& 16 & - & 165.42 \\
& \multicolumn{1}{l}{{Spiking-Diffusion~\cite{liu2023spiking}}} &{SNN} & {-}& {16} & {-} & {91.98} \\          
& \textbf{SDDPM} &SNN&63.61& 4 & - & \textbf{21.38} 
\\ \midrule
\multirow{4}{*}{CelebA$^*$} 
& DDPM~\cite{ho2020denoising} & ANN & 64.47 & / & -& 20.34\\
\cmidrule{2-7}
 & FSVAE~\cite{kamata2022fully} &SNN& 6.37 & 16 & 3.697 & 101.60          \\ 
 & SGAD~\cite{feng2023sgad} &SNN&-& 16 & - & 151.36 \\        
 & \textbf{SDDPM} &SNN & 63.61 & 4 &  - & \textbf{25.09}\\ \midrule
 \multirow{2}{*}{{LSUN bedroom$^*$}} 
& {DDPM~\cite{ho2020denoising}} & {ANN} & {64.47} & {/} & {-}& {29.48}\\
\cmidrule{2-7}      
 & {\textbf{SDDPM}} &{SNN} & {63.61} & {4} &  {-} & {47.64}\\ \midrule
 \multirow{9}{*}{CIFAR-10}  
& DDPM~\cite{ho2020denoising} & ANN & 64.47 &/ &8.380 &19.04 \\
& DDPM$_\textit{ema}$~\cite{ho2020denoising} & ANN & 64.47 & / &8.846 &13.38 \\
\cmidrule{2-7}
& FSVAE~\cite{kamata2022fully} &SNN & 3.87 & 16 & 2.945 & 175.50   \\
& SGAD~\cite{feng2023sgad} &SNN& - & 16 & - & 181.50 \\
& \multicolumn{1}{l}{{Spiking-Diffusion~\cite{liu2023spiking}}} &{SNN} & {-}& {16} & {-} & {120.50} \\
&\textbf{SDDPM} &SNN&  63.61& 4 & 7.440 & 19.73\\
&\textbf{SDDPM} &SNN& 63.61 & 8 & 7.584 & 17.27\\
&\textbf{SDDPM (TG)} &SNN&  63.61& 4 & 7.482 & 19.20\\
&\textbf{SDDPM (TG)} &SNN& 63.61 & 8 & \textbf{7.655} & \textbf{16.89}\\
\bottomrule
\end{tabular}}
\vspace{-2mm}
\caption{\textbf{Results for different dataset.}
In all datasets, SDDPM (Ours) outperforms all SNN-based baselines and even some ANN models in terms of sample quality, which is mainly measured by FID$\downarrow$ and IS$\uparrow$. Results of $^\triangledown$ and $^\natural$ are taken from \cite{kamata2022fully} and \cite{feng2023sgad}, respectively. $_\textit{ema}$ indicates the utilization of EMA training method~\cite{tarvainen2017mean}. For fair comparisons, we re-evaluate the results of DDPM~\cite{ho2020denoising} using the same U-Net architecture as SDDPM. We employ the symbol `/' to represent `None' since ANN does not have the concept of time step. $^*$ denotes that only FID is used for measurement since these data distributions are far from ImageNet, making Inception Score less meaningful. }
\label{tab:mainresults}
\vspace{-4mm}
\end{table*}

\subsection{Threshold Guidance in SDDPM}
\label{subsec:tg}

Recall that sampling can be achieved by substituting the score $\nabla_{x_t} \log q(x_t)$ with either the score network $s_{\theta}(x_t,t)$ or the scaled noise network -$\frac{\epsilon_{\theta}(x_t,t)}{\sigma(t)}$ while discretizing the reverse SDE as presented in Eq.~\eqref{eq:continuous_back}. 
Because of the inaccuracy of the network estimates, we have the fact that $s_{\theta}(x_t,t) \approx -\frac{\epsilon_{\theta}(x_t,t)}{\sigma(t)} \ne \nabla_{x_t} \log q(x_t)$ in most cases. Therefore, in order to sample better results, we can discretize the following rectified reverse SDE~\cite{kim2022refining}: 
\begin{eqnarray}
\label{eq:rec_continuous_back}
    \textrm{d} x = \left [ f(t)x_t-g^2(t) [s_{\theta}+c_{\theta}](x_t,t)\right ] \textrm{d} t + g(t)\textrm{d} \bar{\omega} ,
\end{eqnarray}
where $s_{\theta}(x_t,t)$ represents the score network or scaled noise network, while $c_{\theta}(x_t,t)=\nabla_{x_t} \log \frac{q(x_t)}{p_{\theta}(x_t,t)}$ denotes the rectified term for the original reverse stochastic differential equation (SDE) with the estimation errors of neural network. The omission of the rectified term $c_{\theta}(x_t,t)$ reduces discretization errors and improves sampling performance. However, the practical calculation of $c_{\theta}(x_t,t)$ presents challenges due to the intractability of $q(x_t)$ and $p_{\theta}(x_t,t)$.

In light of the existence of estimation errors, the formulation in Eq.~\eqref{eq:rec_continuous_back} motivates us to explore if we can improve the sampling performance without additional training by computing $c_{\theta}(x_t,t)$. Although a direct computation of this term is infeasible, we can seek suitable approximations to enhance the effectiveness of our sampling process.
Meanwhile, a crucial parameter in the SNN is the spike threshold  $V_{th}$, which influences the SNN's output. We put forward a threshold guidance (TG) by adjusting the threshold by:
\begin{align}
\label{eq:taylor}
&s_{\theta}(x_t,t,V_{th}') \nonumber \\
&\approx s_{\theta}(x_t,t,V_{th}^0)+\frac{\mathrm{d} s_{\theta}(x_t,t,V_{th})}{\mathrm{d} V_{th}} \mathrm{d} V_{th} + O(\mathrm{d} V_{th}) \nonumber  \\ 
&\approx s_{\theta}|_{V_{th}^0}+ s'_{\theta}|_{V_{th}^0} \mathrm{d} V_{th} + O(\mathrm{d} V_{th}) \nonumber \\
&\approx s_{\theta}(x_t,t)+c_{\theta}(x_t,t) , 
\end{align}
which means that we can adjust the threshold in SNNs to estimate the rectified term $c_{\theta}(x_t,t)$. $V_{th}^0$ represents the threshold utilized during the training stage, while $V_{th}'$ denotes the adjusted threshold employed during the inference stage in Eq.~\eqref{eq:taylor}. The first equation is derived through Taylor expansion. Eq.~\eqref{eq:taylor} indicates that adjusting the threshold can enhance the final sampling outcomes when the derivative term is correlated with the rectified term. Moreover, modifying the threshold allows for the manipulation of both the overall quality and diversity of the generated images, particularly in scenarios where image generation is not highly accurate. A lower threshold encourages the occurrence of more spikes. Experiments show that TG can improve sample quality without extra training. We label cases with decreasing thresholds as inhibitory guidance and the opposite as excitatory guidance.

\begin{figure*}[t]
	\setlength{\tabcolsep}{1.0pt}
	\centering
	\begin{tabular}{c}
  \includegraphics[width=0.24\textwidth]{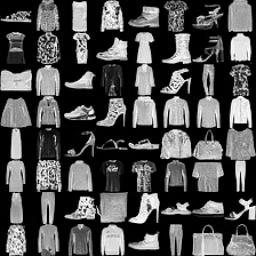}
  \includegraphics[width=0.24\textwidth]{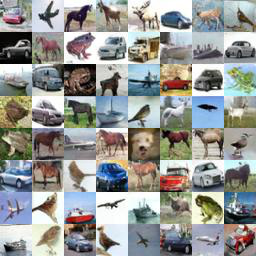} 
  \includegraphics[width=0.24\textwidth]{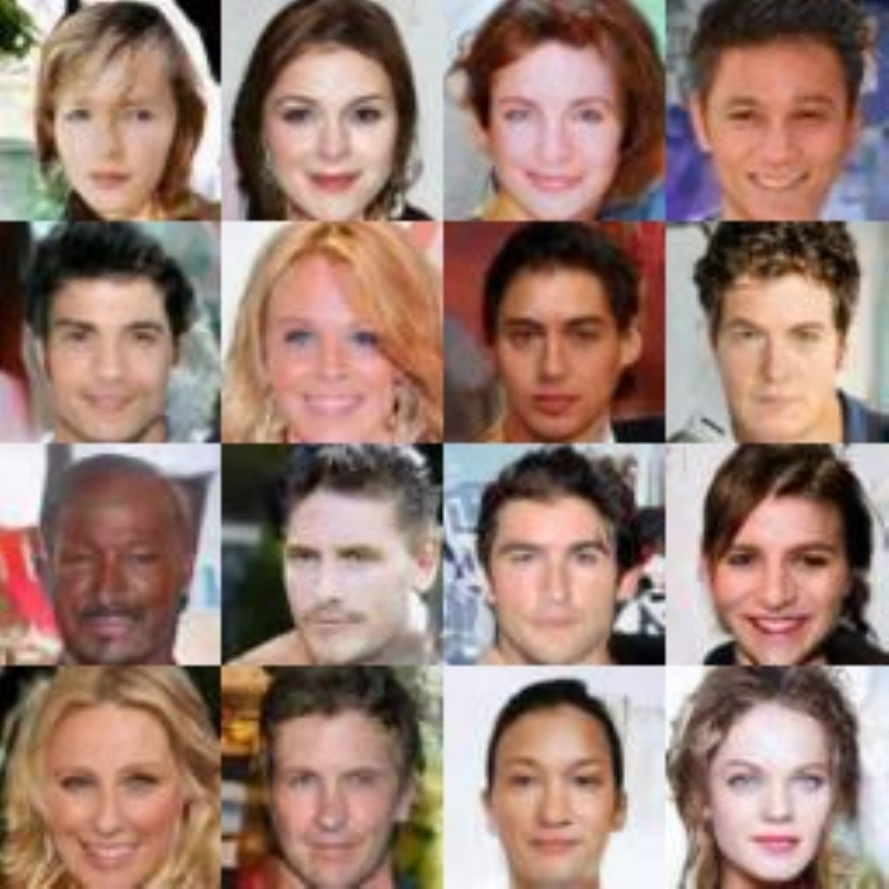}
  \includegraphics[width=0.24\textwidth]{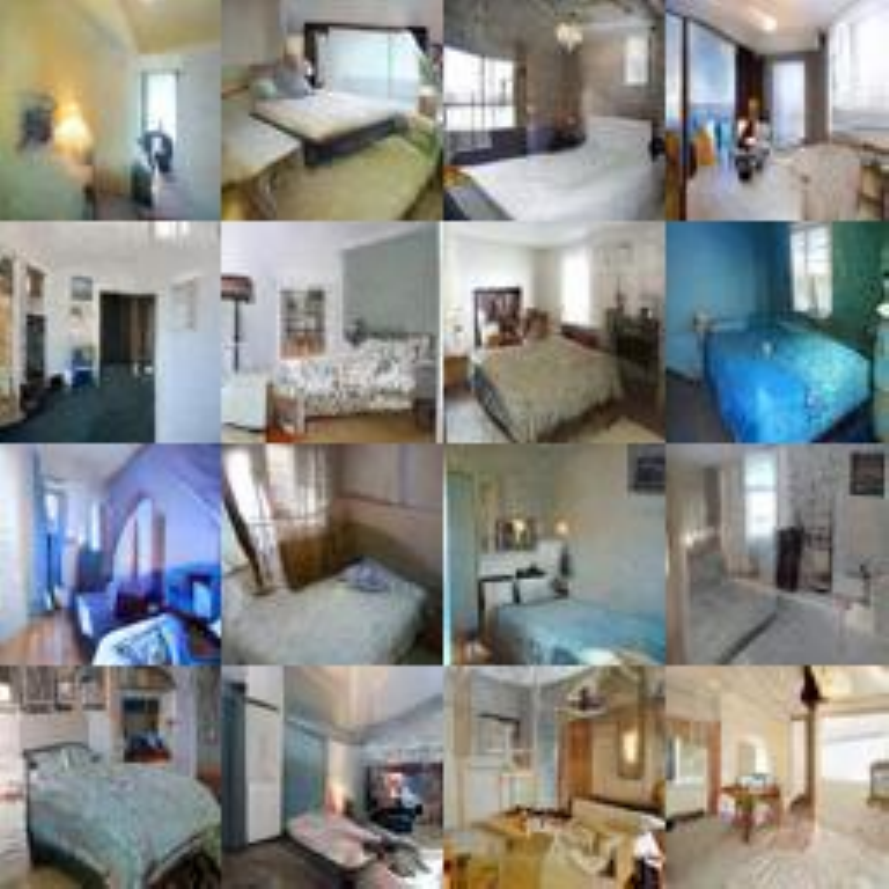}
	\end{tabular}
	\caption{\textbf{Unconditional image generation results on Fashion-MNIST, CIFAR-10, CelebA and {LSUN bedroom} by using SDDPM.}}
	\label{fig:visulization}
	\vspace{-0.35cm}
\end{figure*}

\section{Experiment}

\subsection{Experiment Settings}
\noindent\textbf{Datasets and Baselines}
To demonstrate the effectiveness and efficiency of the proposed algorithm, we conduct experiments on 32$\times$32 MNIST~\cite{lecun2010mnist}, 32$\times$32 Fashion-MNIST~\cite{xiao2017fashion}, 32$\times$32 CIFAR-10~\cite{krizhevsky2009learning}, 64$\times$64 CelebA~\cite{liu2015deep} {and 64$\times$64 LSUN bedroom~\cite{yu2015lsun}.} 
We use existing spiking generative models FSVAE~\cite{kamata2022fully}, SGAD~\cite{feng2023sgad} and {Spiking-Diffusion~\cite{liu2023spiking}} as our baselines. We also compare our results with ANN baselines.

\noindent\textbf{Evaluation Metrics.}
The qualitative results are compared according to Fr\'{e}chet Inception Distance (FID~\cite{heusel2017gans}, lower is better) and Inception Score (IS~\cite{salimans2016improved}, higher is better). {IS evaluates the quality of synthetic images by maximizing the average entropy of Inception V3 model's probability distribution. FID computes the KL divergence between the assumed Gaussian latent spaces of real and generated images. Both metrics are calculated using 50,000 generated images.}

\noindent\textbf{Implementation Details.} 
Our Spiking U-Net inherits the standard U-Net~\cite{ronneberger2015u} architecture and
no attention blocks are used. 
For the hyper-parameter settings, we set the decay rate $e^{\frac{1}{\tau}}$ in Eq.~\eqref{eq:dis_lif1} as 1.0 and the spiking threshold $\vartheta_{\textrm{th}}$ as 1.0. The SNN simulation time step is 4/8. The learning rate is set as 1e-5 with batch size 128 and we train the model
without exponential moving average (EMA~\cite{tarvainen2017mean}). U-Net also does not employ attention blocks, and its training process is consistent with Spiking U-Net. 
More details of the model and the implementation codes can be found in the Supplementary Material.

\subsection{Comparisons with the state-of-the-art}

In Tab.~\ref{tab:mainresults}, we present a comparative analysis of our Spiking Denoising Diffusion Probabilistic Models (SDDPM) with state-of-the-art generative models in unconditional generations. To ensure a comprehensive comparison, we also include results derived from ANNs as benchmarks. Fig.~\ref{fig:visulization} provides a visual representation of the qualitative results obtained. Our results demonstrate that \textit{SDDPM outperforms SNN baselines across all datasets by a significant margin}, even with smaller spiking simulation steps (4/8). In particular, on the CelebA dataset, SDDPM has 4$\times$ and 6$\times$ FID improvement in comparison to FSVAE and SGAD, respectively. Both of these competing models require 16 time steps. On the CIFAR-10 dataset, the enhancement factor is even more substantial, with SDDPM achieving 11$\times$, 12$\times$ and {7$\times$} improvements over FSVAE, SGAD and {Spiking-Diffusion}, respectively. Moreover, the quality of generated samples escalates with an increase in the number of time steps. In specific, our SDDPM attains a level of sample quality that is comparable to the ANN benchmarks with the same U-Net architecture. In certain instances, such as an FID score comparison of 17.27 (SDDPM) against 19.04 (DDPM), \textit{the SDDPM even outperforms the ANN models}. This outcome highlights the superior expressive capability of SNNs employed in our model. {\textit{SDDPM also demonstrates the generative ability on large-scale datasets.} On the LSUN bedroom dataset, which contains more than 3 million images, SDDPM demonstrates commendable qualitative results as depicted in Fig.~\ref{fig:visulization}. }

\subsection{Effectiveness of Threshold Guidance}
In Sec.~\ref{subsec:tg}, we propose a training-free method: Threshold Guidance (TG), which could improve the quality of the generated images by simply changing the threshold of the spiking neurons slightly during inference. As illustrated in Tab.~\ref{tab:TG}, inhibitory guidance helps to further improve the quality of the generated images on two metrics: the Fr\'{e}chet Inception Distance drops from 19.73 to 19.20 upon decreasing the threshold by 0.3\% and the IS score increased from 7.44 to 7.55 upon reducing the threshold by 0.2\%. 
On the other hand, the excitatory guidance also improves sampling quality in some conditions.
Those quantitative results suggest that threshold guidance can provide an effective boost to model performance after the model has been trained, while not costing additional training resources. 

\begin{table}[t]
        \setlength\tabcolsep{15pt}
        \centering
        \resizebox{1\linewidth}{!}{
        \begin{tabular}{cccc}
        \toprule 
    	 \textbf{Method} &\textbf{Threshold} & \textbf{FID}$\downarrow$ & \textbf{IS}$\uparrow$   \\
        \midrule
        Baseline& 1.000   & 19.73 &  7.44  \\
                            \midrule
        \multirow{3}{*}{\shortstack{Inhibitory\\ Guidance}}         & 0.999  & \textbf{\textcolor{blue}{19.25}}  & \textbf{\textcolor{blue}{7.48}}   \\
                            & 0.998  & 19.38   & \textbf{\textcolor{red}{7.55}}  \\
                             & 0.997  & \textbf{\textcolor{red}{19.20}}  &  7.47  \\
                             \midrule
        \multirow{3}{*}{\shortstack{Excitatory\\ Guidance}} & 1.001  & 20.00  &  7.47 \\
                             & 1.002  & 19.98  &  \textbf{\textcolor{blue}{7.48}} \\
                             & 1.003  & 20.04   & 7.46  \\ 
                            \bottomrule
        \end{tabular}}
        \caption{\textbf{Results on CIFAR-10 by different threshold guidances.}  
        The top-1 and top-2 results are colored in red and blue, respectively. The findings indicate that TG can further enhance the FID score by adjusting the spike threshold.}
        \label{tab:TG}
        \vspace{-5pt}
\end{table}

\subsection{Evaluation of the Computational Cost}

To further emphasize the low-energy nature of our SDDPM, we perform a comparative analysis of the FID and energy consumption between the proposed SDDPM and its corresponding ANN model. As shown in Tab.~\ref{tab:energy}, when the (spiking) time step is set at 4, the SDDPM presents significantly lower energy consumption, amounting to merely 37.5\% of that exhibited by the ANN model. Moreover, the FID of SDDPM also improved by 0.47, indicating that our model can effectively minimize energy consumption while maintaining competitive performance. As time steps grow from 4 to 8, we witness a corresponding decline in the FID at the cost of elevated energy consumption. This observation points to a trade-off between FID improvement and the associated energy expenses as time steps increase.

\begin{table}[t]
\setlength\tabcolsep{3.5pt} 
\centering
\begin{tabular}{cccc}
\toprule
\textbf{Models}      & DDPM-ANN & SDDPM-4T       & SDDPM-8T       \\ \midrule
\textbf{FID$\downarrow$}         & 19.04    & 19.20          & \textbf{16.89} \\
\textbf{Energy (mJ)$\downarrow$} & 29.23    & \textbf{10.97} & 22.96  \\ \bottomrule       
\end{tabular}
        \caption{\textbf{Comparisons of energy and FID of SNN and ANN models.} In comparison to ANN, SNN models exhibit reduced energy consumption while attaining superior FID outcomes.}
        \vspace{-5pt}
	\label{tab:energy}
\end{table}

\subsection{Ablation Study} 
\label{subsec:ablation_study}

\noindent\textbf{Impact of different residual learnings.} 
To showcase the superiority of the pre-spike learning approach we utilize, we compare the FID score with that of the traditional spiking residual block on CIFAR-10 dataset. The results of our study, presented in Tab.~\ref{tab:spike_res}, reveal that our pre-spike-based model outperforms its traditional counterpart in terms of FID score, thus demonstrating the supremacy of the pre-spike learning approach within the context of SNN.

\vspace{3pt}\noindent\textbf{Effectiveness of TG on different time steps.} Another critical aspect of our study concerns the examination of our TG strategy's effectiveness with varying spiking time steps. As demonstrated in Tab.~\ref{tab:ablation}, our observations confirm a correlation between an increasing number of time steps and an improvement in the performance of SDDPM. Moreover, the implementation of the TG strategy further amplifies this improvement. For instance, with the application of TG, the FID score improves from 19.73 to 19.20, indicating a relative enhancement of approximately 2.69\%. This improvement suggests that the TG strategy is a significant contributing factor to the overall performance of our SNN model. It is also worth noting that there is an additional enhancement of the FID performance by further refining the TG strategy and increasing the time steps.

\begin{table}[t]
        \centering
        \setlength\tabcolsep{18pt}
        \renewcommand\arraystretch{1.15}
        \resizebox{1\linewidth}{!}{
        \begin{tabular}{c|cc}
        \hline
    	 \textbf{Method} & \textbf{IS}$\uparrow$ &\textbf{FID}$\downarrow$   \\
        \hline
        SNN Resblock
        & 6.25    & 48.69  \\
        Pre-Spike Resblock & \textbf{7.44}  & \textbf{19.73}   \\
                            \hline
        \end{tabular}
        }
        \caption{\textbf{Ablation study on spiking resblock structures.} We evaluate the performances of two SNN residual methods on the CIFAR-10 dataset. The results demonstrate the superiority of the pre-spike residual method.}
	\label{tab:spike_res}
\end{table}

\begin{table}[t]
        \renewcommand\arraystretch{1.15}
        \centering
        \vspace{-3pt}
        \resizebox{1\linewidth}{!}{
        \begin{tabular}{ccc|cc}
        \hline
    	 \textbf{Method} & \textbf{Time Steps} & \textbf{TG} & \textbf{FID}$\downarrow$ & $\Delta$ (\%)   \\
        \hline
        \multirow{4}{*}{\shortstack{SDDPM}}
        & 4    &  & \multicolumn{1}{l}{19.73}&  \multicolumn{1}{l}{+0.00} \\
        &  4   &  \checkmark& \multicolumn{1}{l}{\textbf{19.20 (-0.53)}}  & \multicolumn{1}{l}{+2.69}\\ \cline{2-5}
        &  8 &   & \multicolumn{1}{l}{17.27} & \multicolumn{1}{l}{+0.00} \\
        &  8   & \checkmark& \multicolumn{1}{l}{\textbf{16.89 (-0.38)}}  & \multicolumn{1}{l}{+2.20}  \\ 
        \hline
        \end{tabular}}
        \caption{\textbf{Ablation study on proposed TG and time step.} The experiments are conducted on SDDPM with 1k denoising steps. $\Delta$  represents the improvement of FID. The performance of SDDPM is enhanced by both TG and the increasing time steps.  }
        \vspace{-7pt}
	\label{tab:ablation}
\end{table}

\vspace{-8pt}

\section{Discussion}
\vspace{-2pt}
SDDPM presents a promising opportunity for developing SNN-based generative models, owing to its high-quality generation capabilities. 
Nonetheless, one limitation of our study is that we have not examined higher-resolution datasets (\eg, ImageNet~\cite{deng2009imagenet}, LSUN~\cite{yu2015lsun}). Additionally, employing alternative diffusion solvers, such as DDIM~\cite{song2020denoising}, and Analytic-DPM~\cite{bao2022analytic}, merits consideration in an effort to decrease the number of sampling steps. In future research, we plan to {explore SNN generative models with more neural states} and investigate further applications of SDDPM in the generation domain, attempt to combine it with quantization methods for improving model performance and explore the use of distillation learning in terms of sampling methods.
\vspace{-5pt}
\section{Conclusion}
\vspace{-2pt}
In this work, we propose a new class of SNN-based diffusion models named Spiking Denoising Diffusion Probabilistic Models (SDDPM) that combine the energy efficiency of
SNNs with superior generative performance. As a pioneering endeavor employing SNNs on diffusion models, SDDPM provides remarkable advances in generative performances, significantly surpassing existing SNN benchmarks with mere 4 time steps. Moreover, we introduce a purely Spiking U-Net architecture, designed to maximize the inherent energy efficiency of SNNs. The architecture demonstrates the feasibility of matching the performance of its ANN counterpart while simultaneously offering energy savings of up to 62.5\%. Further, we propose an innovative threshold-guidance strategy to further enhance performance without training. This research signifies a vital step forward in the field of SNN generation, paving the way for future exploration and development in this area.

\clearpage

{\small
\bibliographystyle{ieee_fullname}
\bibliography{egbib}
}

\end{document}